\documentclass[]{spie}  

\usepackage{amsmath,amsfonts,amssymb}
\usepackage{graphicx}
\usepackage{multirow}
\usepackage{xcolor}  
\usepackage{float}
\usepackage[colorlinks=true, allcolors=blue]{hyperref}

\title{Pedicle Screw Pairing and Registration for Screw Pose Estimation from Dual C-arm Images Using CAD Models}

\author[a,b,*]{Yehyun Suh}
\author[a]{Lin Li}
\author[a]{Aric Plumley}
\author[a]{Chaochao Zhou}
\author[b]{Daniel Moyer}
\author[a]{Kongbin Kang}
\affil[a]{AIX Research, Alphatec Spine, Carlsbad, CA, USA}
\affil[b]{Department of Computer Science, Vanderbilt University, Nashville, TN, USA}

\authorinfo{* Corresponding Author: Yehyun Suh, yehyun.suh@vanderbilt.edu}

\begin{document} 
\maketitle

\begin{abstract}
Accurate matching of pedicle screws in both anteroposterior (AP) and lateral (LAT) images is critical for successful spinal decompression and stabilization during surgery. However, establishing screw correspondence, especially in LAT views, remains a significant clinical challenge. This paper introduces a  method to address pedicle screw correspondence and pose estimation from dual C-arm images. By comparing screw combinations, the approach demonstrates consistent accuracy in both pairing and registration tasks. The method also employs 2D-3D alignment with screw CAD 3D models to accurately pair and estimate screw pose from dual views. Our results show that the correct screw combination consistently outperforms incorrect pairings across all test cases, even prior to registration. After registration, the correct combination further enhances alignment between projections and images, significantly reducing projection error. This approach shows promise for improving surgical outcomes in spinal procedures by providing reliable feedback on screw positioning.
\end{abstract}

\keywords{Pedicle Screws, Dual C-arm Images, Screw Correspondence Classification, Screw Pose Estimation}

\section{Introduction}
In spinal or orthopedic surgeries, such as spinal fusion, pedicle screws are inserted into the vertebrae to stabilize the spine and promote healing \cite{Reisener_Pumberger_Shue_Girardi_Hughes_2020, Kosmopoulos_Schizas_2007}. Accurate screw position estimation is crucial for computer-aided rod bending systems and to ensure optimal therapeutic outcomes.  Incorrect positioning can lead to complications such as nerve compression or vertebral instability, which may cause impaired mobility, the need for revision surgeries, or chronic pain syndromes, radiculopathy, and even hardware failure over time \cite{pugely2023modular, gelalis2012loading}. Thus, precise pose estimation is performed using imaging modalities like optical tracking, X-rays, CT scans, or C-arm fluoroscopy to ensure optimal outcomes \cite{alan2024spine}. 

Previous studies have explored screw correspondence classification and registration using various methods. Esfandiari et al. \cite{Esfandiari_Newell_Anglin_Street_Hodgson_2018} utilize epipolar geometry to pair projected screws in bi-planar views . While mostly effective, this approach can fail when two different screws have the similar Euclidean distances along the epipolar lines, leading to potential misclassifications. On the registration side, Uneri et al. \cite{Uneri_DeSilva_Stayman_Kleinszig_Vogt_Khanna_Gokaslan_Wolinsky_Siewerdsen_2015c}employ known-component 3D–2D registration to align screw CAD models with vertebrae using digital radiography; however, this method requires patient preoperative CT, which is not widely available in spine surgeries. The Bendini system (Nuvasive, San Diego, CA, USA) \cite{scholl2020systems} can estimate accurate screw tower pose, but it does not utilize CAD models and instead relies on multiple external devices and optical tracking, and demand calibration procedures between patient and instruments — factors that collectively increase intra-operative complexity. 

This paper introduces a method for screw correspondence classification and pose estimation in dual C-arm images using a screw CAD model. By leveraging the bi-planar views provided by dual C-arm imaging with the CAD model, our approach classifies the correct screw pairings between AP and LAT views and then uses these correspondences to estimate and optimize the screw pose. The enhanced spatial information from the dual views allows for precise 3D pose estimation, while the correct classification of screw pairs ensures that the registration process accurately aligns the CAD model with the real screw positions. This combined method minimizes projection errors and improves alignment, contributing to safer and more effective surgical outcomes.

\begin{figure}[]
    \centering
    \includegraphics[width=\linewidth]{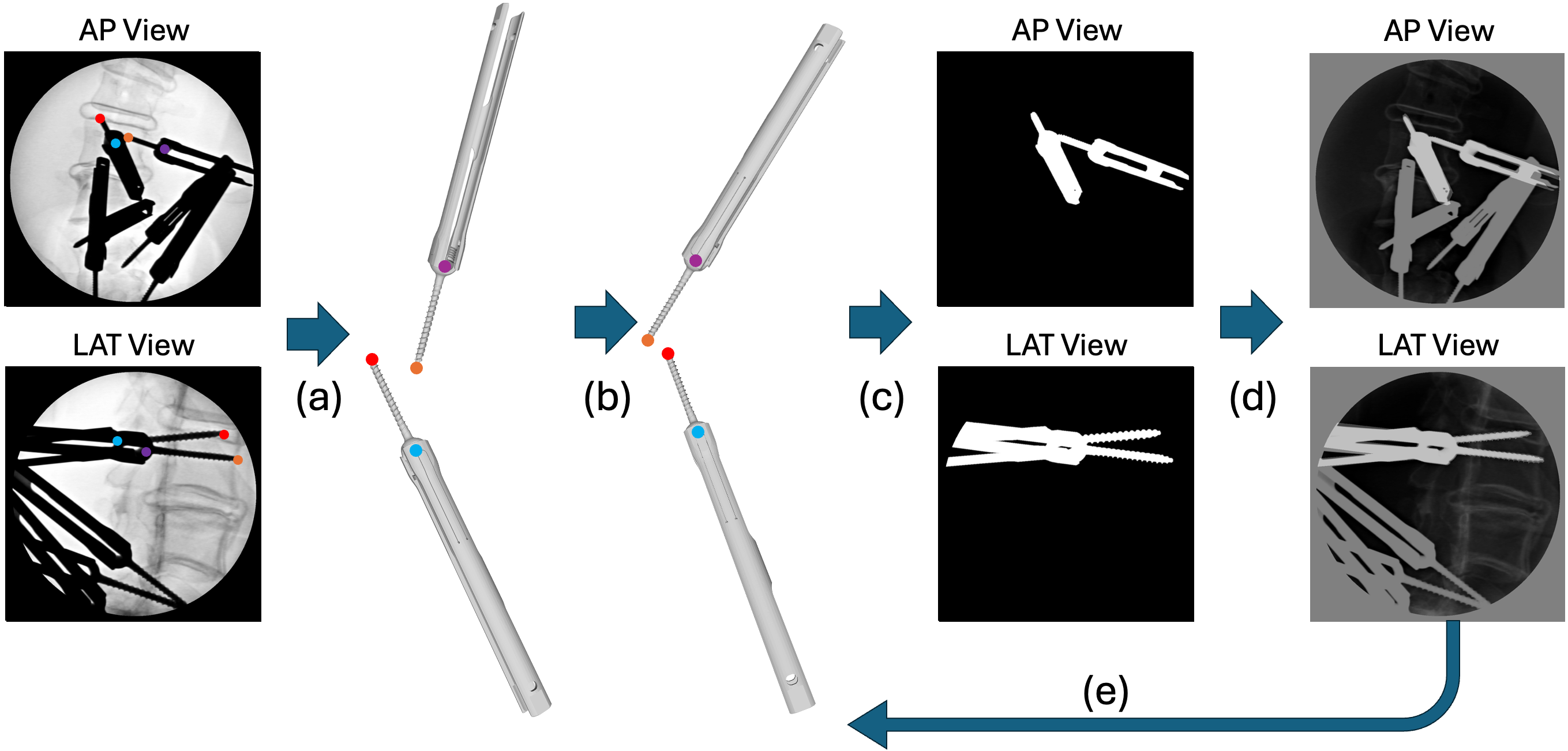}
    \caption{Screw classification and registration pipeline. Red and orange circles represents the screw tip and light blue and purple circles represents the center of screw body for different screws. (a) Calculate 3D coordinates from 2D coordinates in AP and LAT views and align two CAD models. (b) Rotate and translate the models. (c) Project the models into AP and LAT views. (d) Compute the loss between the projection and the real data. (e) Predict the pairs of the screws and repeat the process until the loss converges for accurate screw registration.}
    \label{fig:method}
\end{figure}
\section{Method}
\subsection{Projection of Dual C-arm Views (AP and LAT) and Alignment of CAD Model}
In this study, dual C-arm imaging is employed to provide two distinct views, as illustrated in Figure~\ref{fig:method}. Since pose estimation can be formulated as an optimization problem that minimizes the difference between a target image and a transformed moving image, initializing the pose closer to the true solution---as opposed to a zero initialization---is known to enhance accuracy~\cite{suh2025betterposeinitializationfast}. The dual-view setup enables the estimation of the 3D real-world coordinates for both the tip and the center of the screw's initial pose. These coordinates are derived by manually annotating corresponding landmarks on the 2D projections obtained from both views. The projection matrices \( P^{AP} \) for the AP view and \( P^{LAT} \) for the LAT view are used to map the estimated 3D coordinates onto the respective 2D image planes.

The 3D coordinates $(X_T, Y_T, Z_T)$ and $(X_C, Y_C, Z_C)$ of the screws' tip and center are computed by solving the following equations using least square \cite{Grossmann_Santos_Victor_2005}:
\begin{align}
&X(p_{11}^{AP} - u_{AP}p_{31}^{AP}) + Y(p_{12}^{AP} - u_{AP}p_{32}^{AP}) + Z(p_{13}^{AP} - u_{AP}p_{33}^{AP}) = u_{AP}p_{34}^{AP} - p_{14}^{AP} \nonumber\\
&X(p_{21}^{AP} - v_{AP}p_{31}^{AP}) + Y(p_{22}^{AP} - v_{AP}p_{32}^{AP}) + Z(p_{23}^{AP} - v_{AP}p_{33}^{AP}) = v_{AP}p_{34}^{AP} - p_{24}^{AP} \nonumber\\
&X(p_{11}^{LAT} - u_{LAT}p_{31}^{LAT}) + Y(p_{12}^{LAT} - u_{LAT}p_{32}^{LAT}) + Z(p_{13}^{LAT} - u_{LAT}p_{33}^{LAT}) = u_{LAT}p_{34}^{LAT} - p_{14}^{LAT} \nonumber\\
&X(p_{21}^{LAT} - v_{LAT}p_{31}^{LAT}) + Y(p_{22}^{LAT} - v_{LAT}p_{32}^{LAT}) + Z(p_{23}^{LAT} - v_{LAT}p_{33}^{LAT}) = v_{LAT}p_{34}^{LAT} - p_{24}^{LAT} \nonumber
\end{align}
Here, $(u_{AP}, v_{AP})$ and $(u_{LAT}, v_{LAT})$ represent the tip and center coordinates of the screws on each 2D image planes. The elements \( p_{ij}^{AP} \) and \( p_{ij}^{LAT} \) are part of the projection matrices extracted from the c-arm that map the 3D coordinates into the 2D space.

After estimation of 3D coordinates of the screws' tip and center, the vector \( \mathbf{V} \) that represents the orientation of the screw center to the tip is computed as:
\begin{align}
\mathbf{V} = \mathbf{C}-\mathbf{T}= (X_C - X_T, Y_C - Y_T, Z_C - Z_T) \nonumber
\end{align}
where \( \mathbf{C} = (X_C, Y_C, Z_C) \) is the center and \( \mathbf{T} = (X_T, Y_T, Z_T) \) is the tip of the screws. This vector \( \mathbf{V} \) is then used to align the CAD model of the screw in 3D space by positioning its center and the tip at the corresponding real-world coordinates, as shown in Figure \ref{fig:method}(a).

\subsection{CAD Model Projection}
\begin{figure}[]
    \centering
    \includegraphics[width=\linewidth]{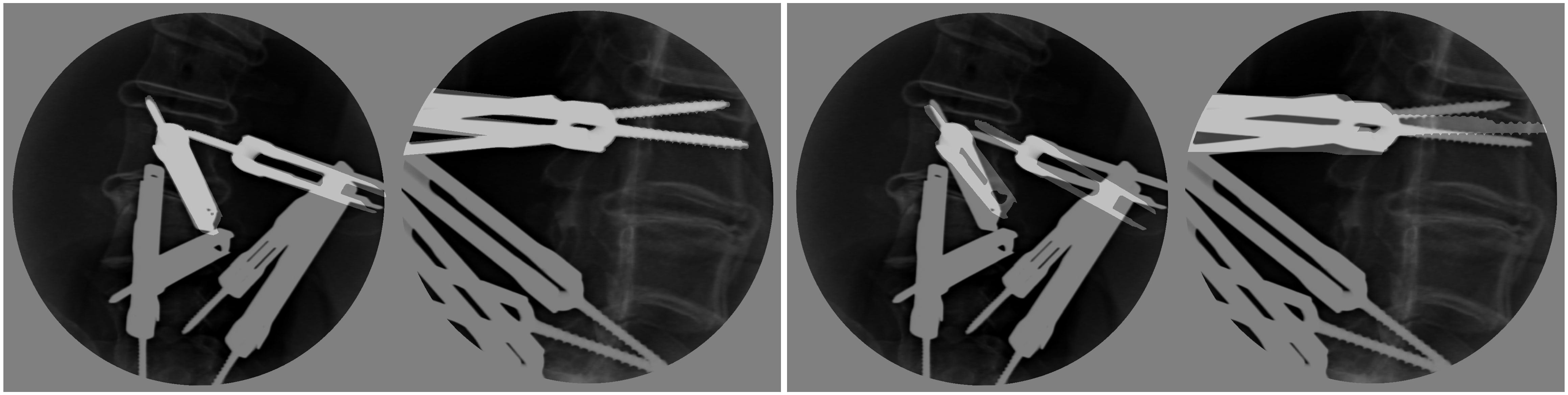}
    \caption{Overlay after after alignment based on 2D coordinates and rotation on axial axis. Image on the left shows the result from the correct combination and image on the right shows the result from the wrong combination. The screw in the white color is the projection from the 3D CAD model and the screw in the gray color is the real screw.}
    \label{fig:overlay after axial rotation}
\end{figure}

Once the CAD models are aligned, we rotate and translate the models to look for the optimal transformation, as shown in Figure \ref{fig:method}(b). Then, the models are projected back onto the AP and LAT views using the respective projection matrices, as demonstrated in Figure \ref{fig:method}(c). These projections provide a 2D binary map in both views and the maps are created from projecting the meshes, covering the entire projection region. Let $X_1$, $X_2$, and $X_3$ represent the 2D coordinates of the three vertices projected from triangle in the CAD model on to the 2D image plane. These coordinates form a triangle  $\triangle (\mathbf{X}_1, \mathbf{X}_2, \mathbf{X}_3)$ and the binary mask \( I_{\text{proj}}(x, y) \) is defined as:
\begin{equation}
I_{\text{proj}}(x, y) = 
\begin{cases} 
1 & \text{if } (x, y) \text{ is inside } \triangle (\mathbf{X}_1, \mathbf{X}_2, \mathbf{X}_3) \\
0 & \text{otherwise} \nonumber
\end{cases}
\end{equation}
Here, for each triangle in the 3D model, the corresponding 2D projection forms a triangle on the 2D image. The pixels that lie within this triangle are assigned a value of 1 in the binary mask, representing the projected region of the screw.

Then, background removal is applied by multiplying the projected image \( I_{\text{proj}}(x, y) \) with a pre-determined background mask \( M_{\text{bg}}(x, y) \), which eliminates irrelevant regions. The final binary map \( I_{\text{final}}(x, y) \) is computed as:
\begin{equation}
I_{\text{final}}(x, y) = I_{\text{proj}}(x, y) \cdot M_{\text{bg}}(x, y) \nonumber
\end{equation}

\subsection{Loss Calculation}
Similarity between the projection and the ground truth image is evaluated using the Gradient Correlation Loss (GCL) \cite{Penney_et_al_GCL}, which measures the alignment of the image gradients, as shown in Figure \ref{fig:method}(d). Given the projected image \( I_{\text{proj}} \) and the real image \( I_{\text{real}} \), the gradients in the \(x\)- and \(y\)-directions are:
\begin{equation}
\nabla I_{\text{proj}} = \left( \frac{\partial I_{\text{proj}}}{\partial x}, \frac{\partial I_{\text{proj}}}{\partial y} \right), \nabla I_{\text{real}} = \left( \frac{\partial I_{\text{real}}}{\partial x}, \frac{\partial I_{\text{real}}}{\partial y} \right) \nonumber
\end{equation}
The Gradient Correlation Loss \( \mathcal{L}_{\text{grad}} \) can be simplified as:
\begin{equation}
\mathcal{L}_{\text{grad}} = -\frac{\sum_{i,j} \nabla I_{\text{proj}}(i,j) \cdot \nabla I_{\text{real}}(i,j)}{\sqrt{\sum_{i,j} \|\nabla I_{\text{proj}}(i,j)\|^2} \cdot \sqrt{\sum_{i,j} \|\nabla I_{\text{real}}(i,j)\|^2}} \nonumber
\end{equation}
Here, \( \nabla I_{\text{proj}}(i,j) \) and \( \nabla I_{\text{real}}(i,j) \) represent the gradients of the projected and real images at each pixel \( (i,j) \), and \( \|\nabla I_{\text{proj}}(i,j)\|^2 \) and $ |\nabla I_{\text{real}}(i,j)\|^2 $ is the squared magnitude of the gradient in each images. The numerator computes the dot product of the gradients, while the denominator normalizes them by their magnitudes, ensuring scale invariance.

This loss is minimized when the gradients of the two images are aligned, minimizing to $-1$ when they are perfectly aligned. It allows optimization process to have more accurate pose estimation of the screws making correlation between the projection and the ground truth to be similar to each other.

\subsection{Optimization Method}
The optimization minimize the total loss \( \mathcal{L}_{\text{total}} \) which is the mean of the loss from AP and the LAT views :
\[
\mathcal{L}_{\text{total}} = \frac{\mathcal{L}_{AP} + \mathcal{L}_{LAT}}{2}
\]
where \( \mathcal{L}_{AP} \) and \( \mathcal{L}_{LAT} \) are the loss from the AP and LAT views. The optimization is formulated as:
\[
\mathbf{\theta}^* = \arg \min_{\mathbf{\theta}} \mathcal{L}_{\text{total}}(\mathbf{\theta})
\]
where \( \mathbf{\theta}^* \) represents the optimal screw pose parameters (translation and rotation) that minimize the total loss and align the projection of the CAD model with the real screw.

For optimization, differential evolution algorithm is employed \cite{Storn_Price_1997}. This global search optimizer is effective in exploring large search spaces and avoiding local minima \cite{Price_Storn_Lampinen_2014, Ahmad_Isa_Lim_Ang_2022}. It iteratively adjusts the pose of the screws, seeking for the best alignment as shown in Figure \ref{fig:method}(e).

\section{Experiment and Result}
\begin{figure}[]
    \centering
    \includegraphics[width=\linewidth]{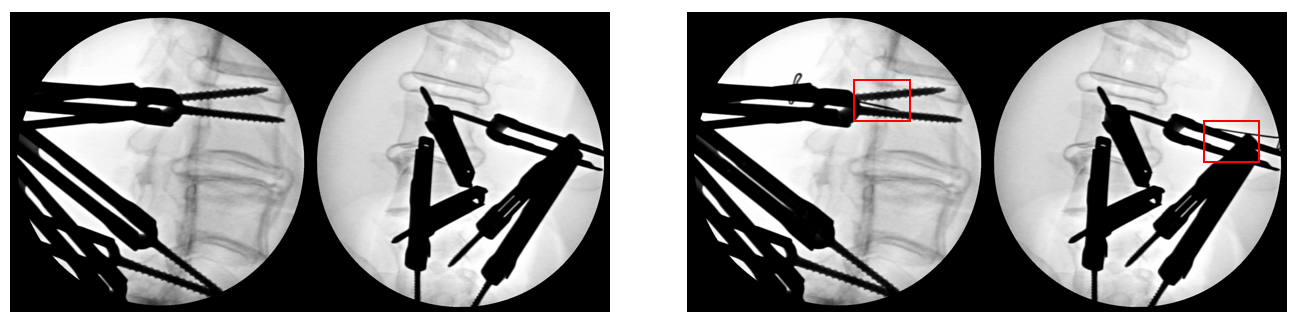}
    \caption{Ground truth screw matching. Left: LAT and AP images of the original screw position. Right: LAT and AP images after placing a needle adjacent to the screw. The needle serves as a spatial reference to establish precise correspondence of the screw across both views.}
    \label{fig:ground truth screw}
\end{figure}

\subsection{Dataset}
To obtain the classification ground truth, we aligned needles along one side of the screw towers. As shown in Figure \ref{fig:ground truth screw}, when capturing images from the AP and LAT views, the screw towers remained aligned with the needle, ensuring consistent matching between the two perspectives. For the classification and registration, we use the images with the needles removed.

\subsection{Screw Correspondence Classification}
\begin{table}[H]
\centering
\begin{tabular}{|c|ccccc|}
\hline
Screw & \multicolumn{1}{c|}{Set 1} & \multicolumn{1}{c|}{Set 2} & \multicolumn{1}{c|}{Set 3} & \multicolumn{1}{c|}{Set 4} & Set 5 \\ \hline
Pre-Registration & \multicolumn{1}{c|}{1} & \multicolumn{1}{c|}{1} & \multicolumn{1}{c|}{1} & \multicolumn{1}{c|}{1} & 1 \\ \hline
Post-Registration & \multicolumn{1}{c|}{1} & \multicolumn{1}{c|}{1} & \multicolumn{1}{c|}{1} & \multicolumn{1}{c|}{1} & 1 \\ \hline
Ground Truth & \multicolumn{1}{c|}{1} & \multicolumn{1}{c|}{1} & \multicolumn{1}{c|}{1} & \multicolumn{1}{c|}{1} & 1 \\ \hline
Accuracy & \multicolumn{5}{c|}{\textbf{100 \%}} \\ \hline
\end{tabular}
\caption{Screw combination correspondence accuracy result. Pre-Registration and Post-Registration stands for classification done before and after the registration. Note that Pre-Registration refers to the state where initial alignment based on 2D landmarks and axial rotation has been applied, prior to full optimization. Ground Truth is the correct combination in each sets.}
\label{table: screw correspondence accuracy}
\end{table}
Accurate screw correspondence between the AP and LAT views is a critical prerequisite for reliable 3D pose estimation in spinal surgeries. Without correctly identifying which screw in one view matches its counterpart in the other, subsequent registration steps may align the wrong models, leading to pose errors and misinterpretation of screw positioning. This is especially important in minimally invasive procedures, where limited visualization increases the risk of misalignment. 

Table~\ref{table: screw correspondence accuracy} summarizes the classification results across all screw sets. The correct combination (Combination 1) was consistently selected both before and after registration, achieving 100\% accuracy in all five test cases. The screw correspondence classification step successfully paired the left and right screws in the AP view with their corresponding screws in the LAT view prior to any registration. 

\begin{figure}[]
    \centering
    \includegraphics[width=\linewidth]{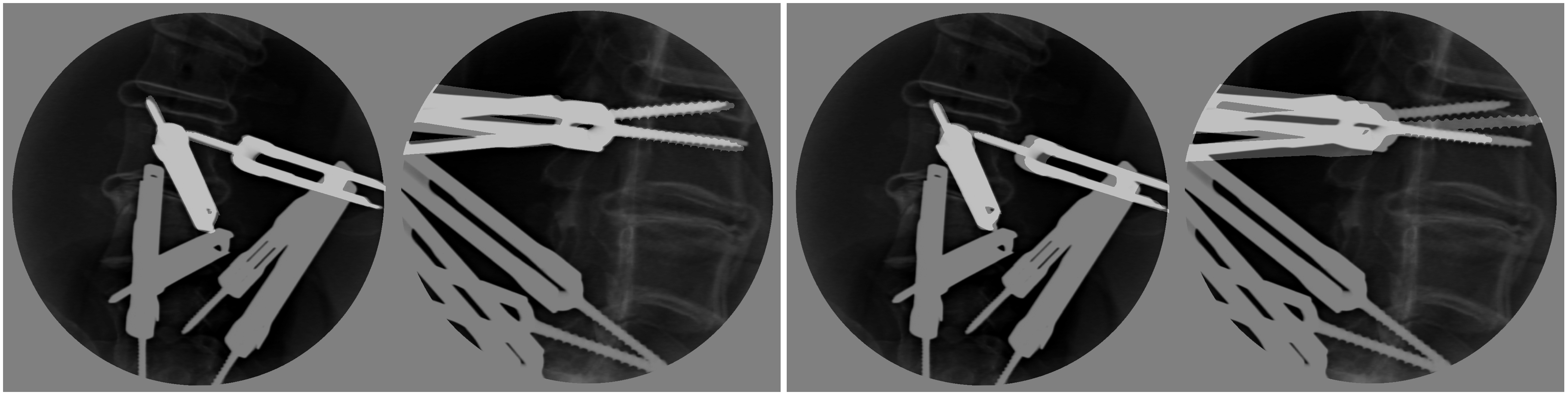}
    \caption{Overlay after registration. Image on the left shows the result from the correct combination and image on the right shows the result from the wrong combination.}
    \label{fig:overlay after registration}
\end{figure}

Quantitative results from Tables~\ref{table: screw correspondence accuracy} and~\ref{table: screw registration} further support this outcome. The combination with the higher Dice score in the screw correspondence stage was selected as the predicted match. These matches with the higher Dice score consistently achieved better results across all metrics in pre-registraion and post-registration scheme. Both the loss values and Dice scores for the correct combination were consistently superior to those of the incorrect one. As shown in Figure~\ref{fig:overlay after registration}, the correct combinations resulted in better visual alignment with the real images compared to the incorrect combinations.

\subsection{Screw Registration}
\begin{table}[H]
\centering
\begin{tabular}{|c|c|cccc|cccc|}
\hline
\multirow{2}{*}{Set} & \multirow{2}{*}{Combination} & \multicolumn{4}{c|}{Pre-Registration} & \multicolumn{4}{c|}{Post-Registration} \\ \cline{3-10} 
 &  & \multicolumn{1}{c|}{Dice} & \multicolumn{1}{c|}{Mean Loss} & \multicolumn{1}{c|}{AP Loss} & LAT Loss & \multicolumn{1}{c|}{Dice} & \multicolumn{1}{c|}{Mean Loss} & \multicolumn{1}{c|}{AP Loss} & LAT Loss \\ \hline
\multirow{2}{*}{1} & 1 & \multicolumn{1}{c|}{\textbf{0.88}} & \multicolumn{1}{c|}{\textbf{-0.24}} & \multicolumn{1}{c|}{-0.23} & -0.24 & \multicolumn{1}{c|}{\textbf{0.89}} & \multicolumn{1}{c|}{\textbf{-0.34}} & \multicolumn{1}{c|}{-0.45} & -0.24 \\ \cline{2-10} 
 & 2 & \multicolumn{1}{c|}{0.66} & \multicolumn{1}{c|}{-0.06} & \multicolumn{1}{c|}{-0.05} & -0.06 & \multicolumn{1}{c|}{0.80} & \multicolumn{1}{c|}{-0.23} & \multicolumn{1}{c|}{-0.34} & -0.11 \\ \hline
\multirow{2}{*}{2} & 1 & \multicolumn{1}{c|}{\textbf{0.84}} & \multicolumn{1}{c|}{\textbf{-0.16}} & \multicolumn{1}{c|}{-0.16} & -0.16 & \multicolumn{1}{c|}{\textbf{0.91}} & \multicolumn{1}{c|}{\textbf{-0.35}} & \multicolumn{1}{c|}{-0.41} & -0.28 \\ \cline{2-10} 
 & 2 & \multicolumn{1}{c|}{0.50} & \multicolumn{1}{c|}{-0.03} & \multicolumn{1}{c|}{-0.04} & -0.01 & \multicolumn{1}{c|}{0.62} & \multicolumn{1}{c|}{-0.18} & \multicolumn{1}{c|}{-0.29} & -0.07 \\ \hline
\multirow{2}{*}{3} & 1 & \multicolumn{1}{c|}{\textbf{0.86}} & \multicolumn{1}{c|}{\textbf{-0.13}} & \multicolumn{1}{c|}{-0.15} & -0.11 & \multicolumn{1}{c|}{\textbf{0.89}} & \multicolumn{1}{c|}{\textbf{-0.29}} & \multicolumn{1}{c|}{-0.18} & -0.39 \\ \cline{2-10} 
 & 2 & \multicolumn{1}{c|}{0.79} & \multicolumn{1}{c|}{-0.09} & \multicolumn{1}{c|}{-0.13} & -0.05 & \multicolumn{1}{c|}{0.76} & \multicolumn{1}{c|}{-0.19} & \multicolumn{1}{c|}{-0.19} & -0.20 \\ \hline
\multirow{2}{*}{4} & 1 & \multicolumn{1}{c|}{\textbf{0.80}} & \multicolumn{1}{c|}{\textbf{-0.07}} & \multicolumn{1}{c|}{-0.08} & -0.07 & \multicolumn{1}{c|}{\textbf{0.83}} & \multicolumn{1}{c|}{\textbf{-0.33}} & \multicolumn{1}{c|}{-0.20} & -0.46 \\ \cline{2-10} 
 & 2 & \multicolumn{1}{c|}{0.60} & \multicolumn{1}{c|}{-0.04} & \multicolumn{1}{c|}{-0.06} & -0.03 & \multicolumn{1}{c|}{0.68} & \multicolumn{1}{c|}{-0.14} & \multicolumn{1}{c|}{-0.17} & -0.11 \\ \hline
\multirow{2}{*}{5} & 1 & \multicolumn{1}{c|}{\textbf{0.77}} & \multicolumn{1}{c|}{\textbf{-0.11}} & \multicolumn{1}{c|}{-0.16} & -0.06 & \multicolumn{1}{c|}{\textbf{0.89}} & \multicolumn{1}{c|}{\textbf{-0.45}} & \multicolumn{1}{c|}{-0.38} & -0.52 \\ \cline{2-10} 
 & 2 & \multicolumn{1}{c|}{0.69} & \multicolumn{1}{c|}{-0.09} & \multicolumn{1}{c|}{-0.15} & -0.03 & \multicolumn{1}{c|}{0.70} & \multicolumn{1}{c|}{-0.19} & \multicolumn{1}{c|}{-0.15} & -0.24 \\ \hline
\end{tabular}
\caption{Quantitative results before and after registration. The left half of the table shows results after alignment based on 2D coordinates and axial rotation. The right half presents results after full 3D registration. AP Loss and LAT Loss refer to the projection losses calculated on each AP and LAT views, respectively. In each screw set, the correct combination consistently outperforms the incorrect one in terms of Dice score and projection loss.}
\label{table: screw registration}
\end{table}

Following screw correspondence classification, registration was performed to determine the optimal screw poses $\theta^*$. As shown in Table~\ref{table: screw registration}, across all test cases, the correct screw pairings consistently achieved higher registration accuracy with lower losses and higher Dice scores when compared to incorrect pairings. These results highlight the critical role of accurate initial correspondence in achieving precise screw pose estimation. The improvements observed from the pre-registration to the post-registration results confirm that proper pairing enhances alignment quality. Notably, even with full optimization, incorrect screw combinations failed to achieve satisfactory registration, indicating that the registration process cannot fully recover from incorrect initial pairings. 

While the correct combination generally outperformed the incorrect one, there was one case in Screw Set 3 where the post-registration AP loss was 0.01 higher for the correct combination. However, the incorrect combination achieved a lower AP loss by overfitting to a single view, while failing to align properly in the LAT view. This highlights the importance of incorporating both views during optimization. Relying on a single projection lacks sufficient spatial information and may lead to misalignment and inaccurate pose estimation—errors that could be critical in a surgical context. 

\section{Conclusion}
In this work, we present a pipeline for pedicle screw correspondence classification and 3D pose estimation using dual C-arm images and CAD models. The method utilizes bi-planar views to manually annotate keypoints, estimate initial 3D coordinates, and align CAD models with the real screws. A projection-based registration process is then applied to refine the screw poses by minimizing a loss function that compares the projected CAD model against the real fluoroscopic images. Experimental results across multiple screw sets demonstrate that the proposed approach reliably identifies the correct screw combinations and achieves accurate pose alignment, leading to improved Dice scores and lower projection losses across both AP and LAT views.

The novelty of this work lies in its integrated use of dual-view fluoroscopic imaging and CAD-based registration to jointly solve the screw pairing and pose estimation problems. Unlike previous approaches that rely on single-view cues or external tracking systems, our method leverages geometric consistency between AP and LAT views to resolve screw correspondence ambiguities and enhance spatial accuracy. The pipeline further demonstrates that accurate initial pairing is essential for successful registration, as incorrect combinations cannot be corrected through optimization alone. These findings emphasize the value of dual-view constraints in clinical scenarios and offer a practical, image-based solution for improving screw localization during spine surgery. 

\begin{figure}[]
    \centering
    \includegraphics[width=\linewidth]{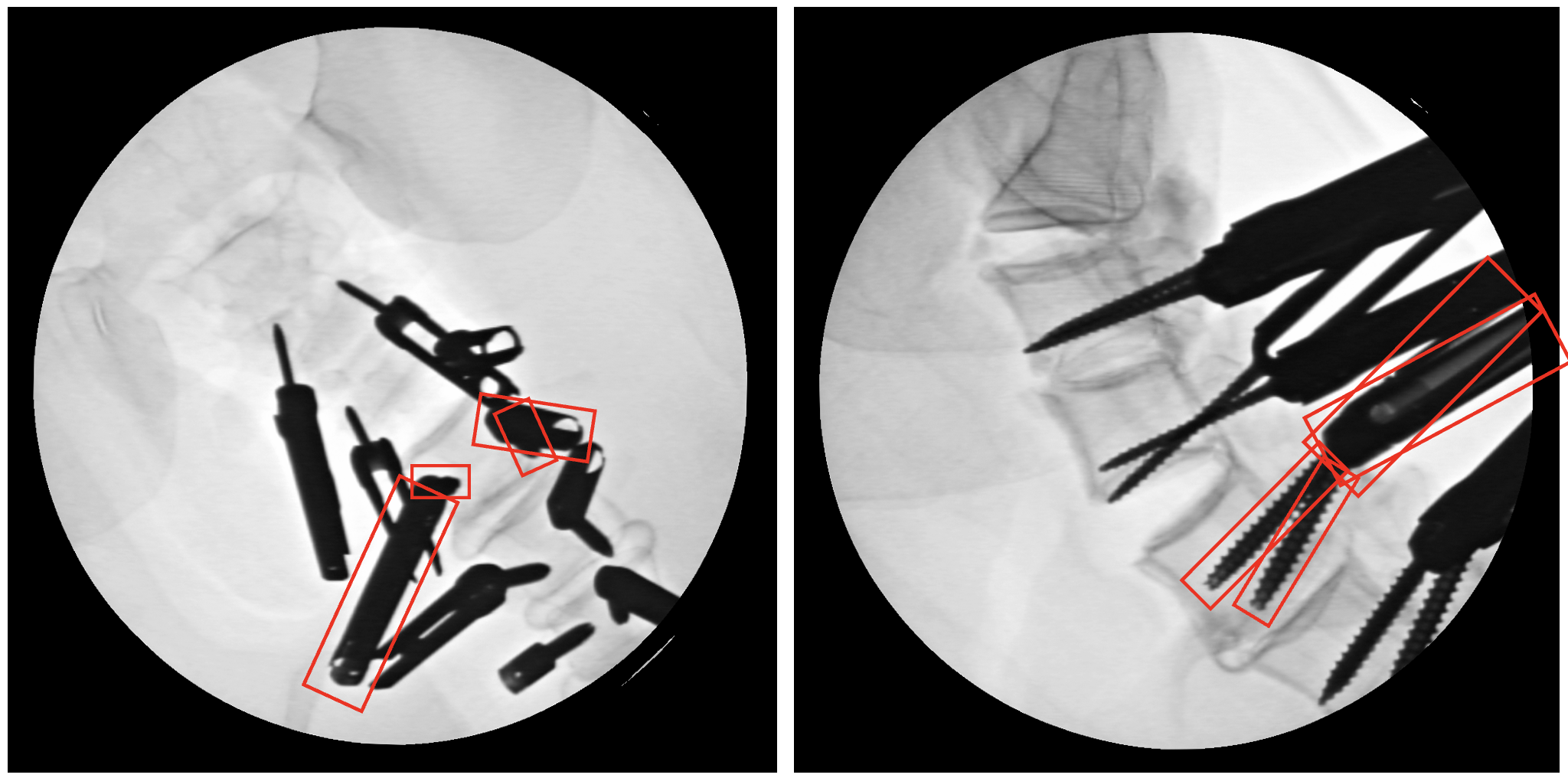}
    \caption{Cases where the screw and tower are not rigidly fixed to each other. Pedicle screws in the red box show differing orientations between the screw and the tower components.}
    \label{fig:24dof}
\end{figure}

\section{Discussion}
While the proposed method shows strong performance in both screw classification and pose registration, there is room for improvement in computational efficiency. The current pipeline applies projection matrix multiplication to the CAD model, which becomes computationally expensive as the complexity of the CAD model or mesh count increases. The use of digitally reconstructed radiographs (DRRs)\cite{gopalakrishnan2022fast,unberath2018deepdrr} instead of direct matrix-based projections may accelerate optimization and improve suitability for intraoperative use.

Another limitation lies in the current modeling of the system with 12 degrees of freedom (DoF), which assumes the screw and tower are rigidly fixed. However, as shown in Figure~\ref{fig:24dof}, the screw and the tower can move independently in practice. Extending the framework to 24 DoF to separately account for each component will be critical for the generalizability of the approach to more complex and realistic surgical scenarios.

Also, we can explore ways to automate specific processes, such as automating the annotation process for the landmarks in the 2D images using deep learning networks \cite{suh2023dilationerosion}. Moreover, the current dataset, while sufficient for initial evaluation, remains limited in scope. Ongoing data collection efforts aim to incorporate a wider variety of screw types, patient anatomies, and imaging conditions. This expanded dataset will support broader validation and increase the method’s generalizability to challenging clinical cases, including overlapping or occluded screw projections.

\acknowledgments 
We would like to acknowledge ATEC AIX System team to assist experiment and E. Zamarripa and T. Semingson for providing ATEC screws and CAD models, as well as B. Aubert from EOS Imaging for discussion.

\bibliography{report} 
\bibliographystyle{spiebib} 

\end{document}